\documentclass[twoside,leqno,twocolumn]{article}

\usepackage[letterpaper]{geometry}

\usepackage{ltexpprt}
\usepackage{hyperref}
\usepackage{graphicx}
\usepackage{diagbox}
\usepackage{graphicx}
\usepackage{algorithm}
\usepackage{algorithmic}
\usepackage{amsmath}
\usepackage{amssymb}
\usepackage{newfloat}
\usepackage{listings}
\usepackage{caption}
\usepackage{subfigure}
\begin{document}

\newcommand\relatedversion{}

\title{\Large Graph Neural Network with Curriculum Learning for Imbalanced Node Classification\relatedversion}

\author{Xiaohe Li\thanks{Aerospace Information Research Institute, Chinese Academy Of Sciences. Email: \{lixiaohe, wanglei002931, fanzd, dengyw\}@aircas.ac.cn}
\and Lijie Wen\thanks{School of Software, Tsinghua University, Beijing, China. Email: wenlj@tsinghua.edu.cn,hxm19@mails.tsinghua.edu.cn} \thanks{Lijie Wen and Yawen Deng are the co-corresponding authors.} 
\and Yawen Deng\footnotemark[1] 
\footnotemark[3]
\and Fuli Feng\thanks{National University of Singapore. Email: fulifeng93@gmail.com}  
\and Xuming Hu\footnotemark[2]
\and Lei Wang\footnotemark[1]
\and Zide Fan\footnotemark[1]
}
\date{}

\maketitle


\fancyfoot[R]{\scriptsize{Copyright \textcopyright\ 2022 by SIAM\\
Unauthorized reproduction of this article is prohibited}}





\begin{abstract} \small\baselineskip=9pt Graph Neural Network (GNN) is an emerging technique for graph-based learning tasks such as node classification. In this work, we reveal the vulnerability of GNN to the imbalance of node labels. Traditional solutions for imbalanced classification (e.g. resampling) are ineffective in node classification without considering the graph structure. Worse still, they may even bring overfitting or underfitting results due to lack of sufficient prior knowledge. To solve these problems, we propose a novel graph neural network framework with curriculum learning (GNN-CL) consisting of two modules. For one thing, we hope to acquire certain reliable interpolation nodes and edges through the novel graph-based oversampling based on smoothness and homophily. For another, we combine graph classification loss and metric learning loss which adjust the distance between different nodes associated with minority class in feature space. Inspired by curriculum learning, we dynamically adjust the weights of different modules during training process to achieve better ability of generalization and discrimination. The proposed framework is evaluated via several widely used graph datasets, showing that our proposed model consistently outperforms the existing state-of-the-art methods.
\end{abstract}

\section{Introduction}
\begin{figure}[tb]
  \centering
  \subfigure{\label{fig1:1}
  \begin{minipage}[t]{0.2\textwidth}
  \centering
  \includegraphics[width=1\textwidth]{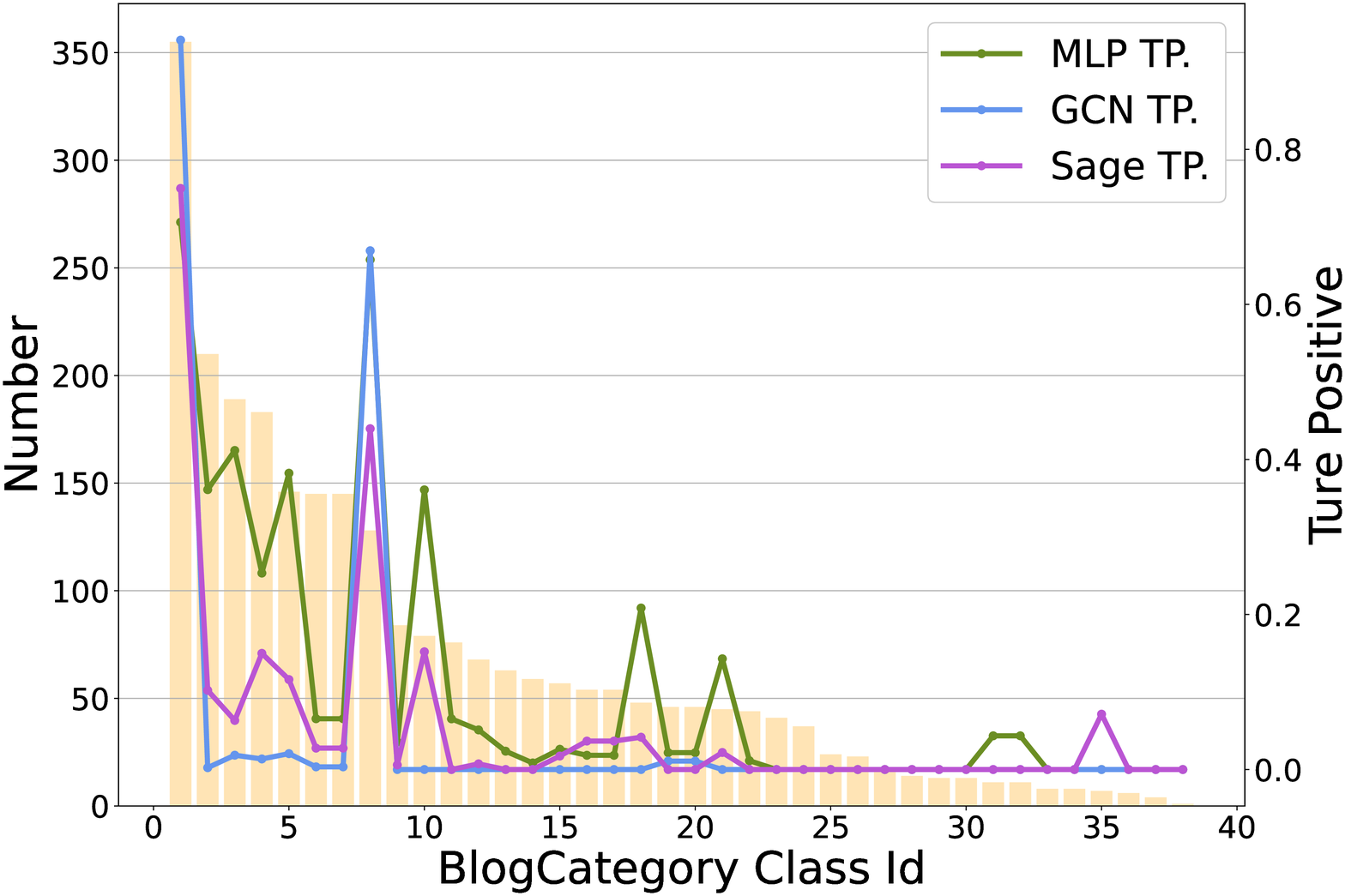}
  \end{minipage}
  }
  \subfigure{\label{fig1:2}
  \begin{minipage}[t]{0.2\textwidth}
  \centering
  \includegraphics[width=1\textwidth]{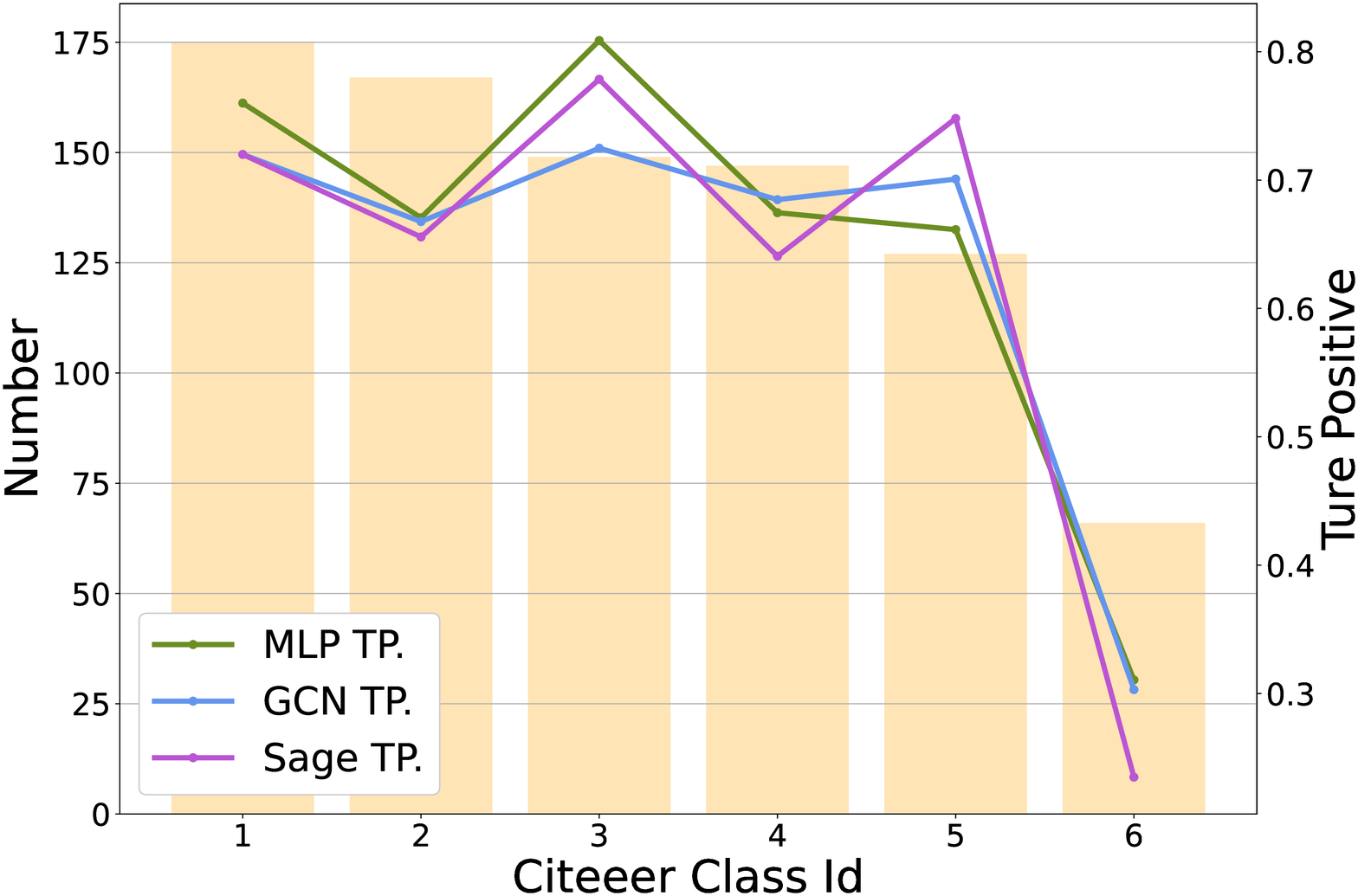}
  \end{minipage}
  }
  \caption{Imbalanced training data class distributions in different class (from majority to minority) on the BlogCategory and Citeeer datasets. 
  It also includes the testing True Positive (TP.) score of GCN\cite{4}, MLP and GraphSAGE\cite{9} on two datasets. We
  can observe a significantly high correlation between the accuracy and class proportion.}
  \label{fig1}
\end{figure}

Graph neural network (GNN), as an novel method to utilize structured data in non-euclidean space, has been widely studied in recent years\cite{1}. GNN can solve many different tasks on complex graphs, such as node classification\cite{4}, edge prediction, clustering and so on. In mainstream machine learning research fields, such as computer vision (CV), researchers usually focus on various problems in semi-supervised classification task. Especially, when there are few labeled samples that can be used for model training and the ratio of each class in the training set is disproportionate.
Similarly, problems exist in the practical applications of graph, for example, 
real-world datasets always have imbalanced class distributions shown in Figure~\ref{fig1}.
At present, the commonly used graph neural network models rely on propagation-aggregation mechanism, such as GCN, GraphSAGE. When these methods meet such imblanced situations, minority samples can not influence others effectively due to insufficient connection edges, meanwhile, data hungry limitation and class imbalanced trouble will decrease the accuracy of the deep graph classifier. 
We verify these phenomenons through some experiments and report the results in Figure~\ref{fig1}.

In some well-studied domains, such as CV, 
there are several kinds of methods to solve the problems of imbalanced class distribution:
1) One of the classic methods is the resampling mechanism including oversampling and downsampling, aiming to balance the data distributions of minority classes and majority classes. Generating new samples by interpolation such as SMOTE\cite{15} comes from a common starting point.
However, it potentially brings negative impact on accuracy in each class for evaluation due to overfitting or underfitting problems since improper resampling scaling can lead to excessive utilization of minority samples or discarding useful information in the majority samples.
2) In addition, there is another kind of method called cost-sensitive learning,
which increases the weight of the minority class classification loss.
In the same way, it is difficult to determine the exact proportions for different classes due to missing priori knowledge of datasets.

While the above mentioned strategies have their pros and cons, we put forward a key argument that they are not directly applicable on the graphs. At present, researches focusing on imbalanced node classification are rare. A few proposed graph-based resampling methods acceptably help to improve the quality of node representations, but it also has some limitations: 1) In the early stage of training,
the quality of the generated embedding is relatively unstable. The features of excess minority samples will be propagated globally, which make the original node representations affected by the infiltration of confusing information.
2) Maintaining a balanced distribution throughout the training process will cause negative effect on generalization since the classifiers in GNN emphasize the minority nodes excessively, especially for overly imbalanced datasets. 
3) It is impossible to utilize the structure information by directly resampling on graph. To enhance the graph structure, Zhao et al.\cite{25} extend previous oversampling algorithms and train an edge generator. But they ignore the complex relationships and feature interactions between nodes, so that the quality of the generated edges cannot be guaranteed. 

In this work, we explore the central theme of solving imbalanced node classification problem. Towards our target, we first propose the novel graph-based oversampling method: adaptive graph oversampling, which supplements the nodes and edges in the graph. 
In particular, we hope to acquire certain reliable interpolation nodes based on the existing embeddings in the model's middle layers. The generated nodes need to be connected with other parts of the graph, therefore we generate new edges based on two essential attributions smoothness and homophily. These synthetic structures help to enhance the reliability of classifier for minority class. Furthermore, besides improving the classifier, we are also committed to improving the representation quality of original and synthetic minority class nodes. We pay attention to metric learning for minority class incremental rectification and add a neighbor-based triplet loss which discovers sparsely boundaries of minority class samples. It looks like the class rectification loss (CRL) function introduced by Chen et al.\cite{26} Based on this intuition, we combine the graph classification loss and the neighbor-based triplet loss which separates different samples associated with the minority class in feature space.

In order to overcome aforementioned overfitting problems and prevent the loss of majority class information, we propose to control the training process from easy to hard inspired by curriculum learning. In this way, the classifier works in the original distribution of the overall graph at the beginning of the training process. Then, nodes and edges are gradually generated to increase the influence of minority class in the graph, which makes classifier focus more on the difficult conditions. Moreover, we believe that 
these two above losses focus on different objectives. The label classification loss is mostly used to correctly assign specific labels,
while the neighbor-based triplet loss mainly optimizes the soft features of minority class by adjusting the distance between nodes.
Similarly to the previous point, we expect that the framework first learns the appropriate feature representations and then generates high-quality samples to correctly optimize the classifier. These two components can be defined by the overall curriculum learning strategy, which leverages the learning process. They have the opposite trend and should be coordinated together.
Since our model is trained with novel curriculum learning, we name it as graph neural network framework with curriculum learning (GNN-CL).
The main contributions of this work are summarized as follows:
\begin{itemize}
\item For the first time, we introduce the curriculum learning idea into graph
classification task which controls the training process from easy to hard. 
Based on these, two components aimed to overcome graph imbalance problem are proposed 
for dynamic sampling operation and loss backward propagation.

\item We propose a novel graph neural network framework with curriculum learning (GNN-CL)
for imbalanced node classification. Further demonstration indicates that these unified methods can significantly promote representation earning and classifier training under the original data distribution.

\item We compare proposed GNN-CL with many state-of-the-art baselines on five
real-world datasets for semi-supervised node classification task to show the effectiveness of node embeddings learned by our model. Further analysis and visualization intuitively reveal the superiority of proposed model.
\end{itemize}

\section{RELATED WORK}
\subsection{Imbalanced Learning}
At present, there are different groups of methods to reduce the bias\cite{10} caused by majority class nodes in the model training process by increasing the importance of minority class nodes. 1) Resampling: transfering the data into a balanced distribution\cite{11}\cite{12}. These methods can be disassembled into two types: one is oversampling, which adjusts the proportion of data samples by simply copying minority class samples. In addition, an advanced sampling method called SMOTE\cite{15} expands artificial samples by interpolating similar samples. The other is undersampling, which balances the sample proportions by abandoning majority classes of samples. However, such methods may cause overfitting or underfitting problems due to repeating visiting duplicated samples or giving up important information. 2) Reweighting: different from the idea of changing the sample set, there is another kind of method to keep the balance of the training process by adjusting the objectives. Cost-sensitive learning intends to assign varying weights to different classes, such as a higher loss for minority class samples\cite{13}\cite{14}. In contrast, the threshold-adjustment technique changes the decision threshold when testing\cite{16}. However, due to the lack of prior knowledge of different datasets and backgrounds, it is difficult to ensure how to set the weight correctly. 3) Hybrid: some methods devote to combining the above categories, for example, EasyEnsemble and BalanceCascade propose a committee of classifiers on undersampled subsets\cite{17}. SMOTEBoost combines the boosting technology and SMOTE oversampling. Furthermore, researchers introduce some novel methods, such as metric learning\cite{18}, meta-learning. And there are also neural network based methods for imbalanced data learning. 
However, few studies have worked on the imbalanced classification problem on graphs.
 
 \subsection{Graph Neural Network}
 Graph neural network (GNN) is a classical model widely used in recent years, which transforms the complicated input graph-structure data into meaningful representations for downstream mining tasks by information passing and aggregation according to dependencies in networks. Among all GNNs, graph convolutional network (GCN) are thought to become a dominating solution, falling into two categories: spectral and spatial methods. As for spectral domains, Bruna et al.\cite{20} proposed to utilize fourier base vector to perform convolution in the spectral domain. ChebNet\cite{19} introduced that smooth filters in spectral convolutions can be well-approximated by K-order Chebyshev polynomials. Kipf et al.\cite{4} presented a convolutional architecture via a localized first-order approximation of spectral graph convolutions which further constrains and simplifies the parameters of ChebNet\cite{19}. On the other hand, spatial methods are defined directly on the graph, operating on the target node and its topological neighbors, so as to realize the aggregation operation on the graph-structure. For example, Hamilton et al.\cite{9} proposed GraphSAGE which generated embeddings by sampling and aggregating features from nodes’ local neighborhood. In addition, there are many works utilizing attention layers in neural networks, such as GAT\cite{21}, which leverages masked self-attention to enable specifying different weights to different nodes in the neighbors. 
 However, these methods do not deal with the bias caused by majority class nodes in the process of implementation, for which they are not suitable for imbalanced node classification problem.

\section{THE PROPOSED MODEL} 
\begin{figure}
\centering
\includegraphics[width=0.48\textwidth]{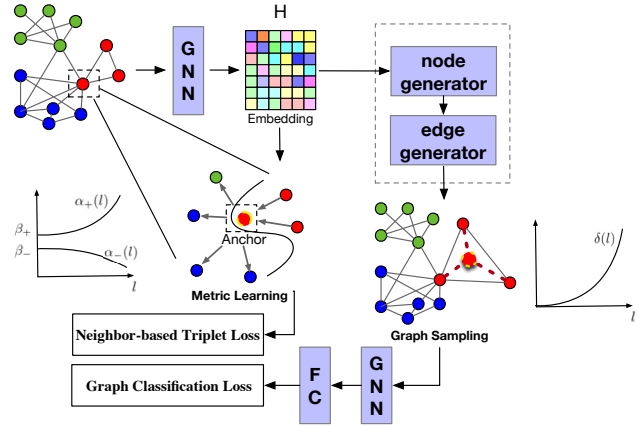}
\caption{The proposed model framework. It includes two loss functions: graph classification loss and neighbor-based triplet loss. They are scheduled through a unified curriculum learning framework. }
\label{model}
\end{figure} 
For the problem of semi-supervised node classification from imbalanced graph data, we hope to construct a graph neural network model with curriculum learning which has the ability to infer the type of unknown nodes. Before detailed introduction, we give some necessary definitions first. In a imbalanced graph, denoted as $G = (V,E,F)$, where $V=\{v_i\}^{N}_{i=1}$ indicates 
the $N$ node set and $E=\{e_{i,j}\}$ indicates the edge set which connect $v_i$ and
 $v_j$. Let $F\in \mathbb{R}^{n\times d}$ denotes
 samples feature matrix and $y\in\{1, 2,..., C\}$ is its corresponding label,
where $C$ is the number of classes. In imbalanced datasets, The number of nodes $|c_{i}|$
available for multi-label tasks varies greatly. 

In order to jointly learn node features and the label
classifiers from class imbalanced training set in an end-to-end process, we propose a novel graph
 neural network framework with curriculum learning for imbalanced node classification problem,
consisting of two novel components shown in Fig~\ref{model}. The first one is an adaptive graph oversampling, of which the key idea is to interpolate the most significant samples related to the original structure. The main purpose is to dynamically make the data distribution in the graph from imbalance to balance. The second one is the 
neighbor-based metric learning. By this way, the distances between nodes and their neighbors are regularized according to pseudo labels, so as to dynamically adjust the position of the embeddings of minority class nodes in feature space. The proposed model balances two losses in the whole learning procedure: 
label classification loss and neighbor-based triplet loss.
Above them, we put up an overall curriculum scheduling strategy consisting of two opposite learning curves. 
In the early stage of the training process, our proposed framework focuses more on optimizing feature propagation and reducing biased noises in the soft feature space. As training goes on, it gradually pays more attention to the average accuracy in each class.

\subsection{Adaptive Graph Oversampling}
As mentioned in the previous section, we firstly need an innovative oversampling strategy to deal with imbalanced graph learning by acquiring an augmented balanced graph reasonably. The SMOTE method proposed by Chawla et al.\cite{15} is one of the most commonly used resampling methods at present,
which is realized by adding synthetic samples between the feature representations of minority class samples.
However, the SMOTE method is not suitable for being used directly since the relationship information contained in the graphs, so that we adjust it based on the characteristics of graph and propose a novel adaptive graph oversampling module consisting of node and edge generators.

As for node generator, we reform the origin SMOTE method by using the $k$-nearest neighbor nodes of the same class in the feature space to guide model to interpolate new minority class nodes.
In particular, if all nodes in the neighbors belong to the same class or different class,
we will ignore such nodes which is similar to the practice in SMOTE-Boardline\cite{22}.
There are two reasons: 1) The features of nodes in the same class are close in the feature space. 
In this case, we use multiple neighbors to construct similar nodes with high reliability.
(2) Selecting the hard minority nodes as sample set help
focus on directly optimizing the classification boundary to improve the stability of the model.
Explicitly, we firstly acquire the middle layer node representations $h_v\in \mathbb{R}^{d}$ of a node $v\in V$ obtained from the general GNN model, 
in which feature and structure information are fused. The formula of $h_v$ in $l^{th}$ layer is as follows: 
\begin{equation}
h_v^{(l)}=GNN^{(l)}(h_v^{(l-1)},({h_{v'}^{(l-1)}:v'\in N(v)})),
\end{equation}
where $N(v)$ denote the set of neighbors of node $v$. Since the same class of nodes usually forms a community in the feature space,
we use $k$-nearest neighbor method to interpret the candidate neighbors in the embedding space for
the feature representations of minority class nodes inspired by SMOTE-$boardline$. We select node $v_i \in V$ from the minority class $C^{-}$ with curriculum probability function $\delta(l)$, where $l$ refers to current training epoch. Then we calculate its $k$-nearest neighbors $KNN(v_i)$ from the whole training set $T$. 
Suppose the number of the same class examples among the above neighbors is $k'$. If $ 0 < k' < k $, $v_i'$ is considered to 
be easy misclassification which we denote as danger node. For each danger sample $v_i'$, we select the sample set $P$ belonging to the same class with $v_i'$ in $k$-nearest neighbors $KNN(v_i)$, and then calculate the differences $D$ using Euclidean distance between $v_i'$ and its neighbors in $P$.
After that, we can generate $|P|$ new synthetic minority nodes $\widehat{v_i}$ with the following interpolation:
\begin{equation}
{h_{\widehat{v_i}}^{(k)}}' = h_{v_i'}^{(k)} + r_j\times D_j,\ s.t.\ 0<r_j<1.
\end{equation}
Here, $r_j(j=1,\cdots,|p|)$ is a random number.
These synthetic nodes obtained through oversampling process
make the proportion of minority class higher in training process.

Next, we introduce the second part: edge generator. In order to effectively apply GNN model, it is necessary to generate new edges for synthetic nodes that can be adapted to the original graph. Hou et al.\cite{23} propose that smoothness and homophily can measure the quality of information obtained from graph data. Inspired by them, we design an indicator to measure the quality of graph structure, so that we can obtain the generated edges via designed edge generator and then gain an augmented edge set $E'$. In this way, the optimized graph is more suitable for the execution of GNN model. To compute the existing probabilities of latent edges related to synthetic nodes for each round, we use the classical attention method to get the coefficients ${a^{(k)}_{i,j}}'$ in round $k$, closely related to the context vector of the node and its neighbors:
\begin{equation}\label{eq3}
{a^{(k)}_{i,j}}' = \frac{exp(\sigma ((W_1^{(k)} h_{v_i})^{T}\cdot (W_1^{(k)} h_{v_i} - W_2^{(k)} h_{v_j})))}{\sum\limits_{v_l\in N(v_i)} exp(\sigma ((W_1^{(k)} h_{v_i})^{T}\cdot (W_1^{(k)} h_{v_i} - W_2^{(k)} h_{v_l})))}.
\end{equation}
Here, $W_1$ and $W_2$ are two learnable matrice.
In Eq.\ref{eq3}, we use the representation difference of node $v_i$ and node $v_j$, which is inspired by the fact that decreasing smoothness meaning that the neighbors
can contribute greater information gain. Finally, we give the loss function $L_{edge}$ for training the edge generator:
\begin{equation}
L_{edge} = \|A'-A\| + \|M'-M\|.
\end{equation}
In the formula, $A$ refers to the factual adjacency matrix and $M$ refers to the homophily matrix. $M_{i,j}$ is 1 means node $v_i$ and node $v_j$ are connected in the training set and they belong to the same class. $M'$ is the predicted homophily matrix for labeled nodes, while $A'$ represents the predicted adjacency matrix for existing nodes. The reason for keeping high homophily is that nodes in the same community tend to have connected edges. A lot of work\cite{23} has proved that homophily is the key to improve the performance of GNN models based on propagation-aggregation mechanism.

We hope that the generated edges can maintain the structural characteristics of the original graph and solve the dilemma of class imbalanced. With the help of edge generator, we next offer the integral complement by adding generated edges into the augmented edge set, which are determined by a threshold $\epsilon$:
\begin{equation}
\widehat{A}_{v',u}=
\begin{cases}
1 ,& \ A'_{v',u} \geq \epsilon, \\
0 ,& \ A'_{v',u} < \epsilon.\\
\end{cases}
\end{equation}
Here $\widehat{A}$ is the adjacency matrix containing new points and edges obtained after sampling, which will be used in the following classifier. In the specific implementation, we can limit the candidate set of node $u$, containing the one-hop neighbors of interpolated set $P$ of $v'$.

According to target curriculum probability function $\delta(l)$ , the minority class samples are re-weighted in different epochs to
confirm inclining to balancing gradually.
Specifically, we adopt another GNN block, appended by a linear layer for node classification as follows:
\begin{equation}
p_{v}=softmax(GNN(h_v,({h_{v'}:v'\in \widehat{N}(v)}))),
\end{equation}
where $\widehat{N}$ represents the augmented neighbor set corresponding to $\widehat{A}$. $p(v)$ is the probability distribution on class labels for node $v$.
Therefore, we give the loss function $L_{node}$ for node classification.
\begin{equation}
L_{node} = -\sum_{v\in \mathcal{\widehat{V}}_{l}}\sum_{c=1}^{C} Y_{v}\lbrack c\rbrack\cdot\log(p_{v}\lbrack c\rbrack),
\end{equation}
where $\widehat{V}_{l}$ is the set of labeled nodes, $Y_{v}$ is the one-hot vector indicates the ground-truth labels of nodes.
Finally, the graph classification loss is defined as:
\begin{equation}
L_{GCL} = L_{node} + \lambda \cdot L_{edge}.
\end{equation}
With the guide of labeled data, we can optimize the model via back propagation and learn the embeddings of nodes.

\subsection{Neighbor-based Metric Learning}
In addition to oversampling, hard mining is also an 
important method to accelerate the convergence speed of the learning process and improve the quality of representation embedding for imbalanced data classification.
Similarly, for the hard nodes in the minority class, we hope that they can avoid the
dominant effect of majority classes on the graph.
Chen et al\cite{24}. add a distance-based regularizer to make nodes receive more useful information from the adjacent nodes and less interference noise from remote nodes in the graph topology.

Therefore, we adopt the somewhat similar metric learning, which is realized with specific loss functions, such as contrastive loss, triplet loss, etc. In graph networks satisfying independent identically distributed, minority class nodes have less chance to have the same class neighbors. The aggregation mechanism of GNN model makes nodes get much confusing information, leading to the decrease of performance and even over-smoothing issue. We observe this phenomena on various datasets, which are shown in Appendix A.

In order to solve the above problems, we use the novel triplet loss function to constrain 
the distances between minority class nodes and neighbors by drawing the same class neighbors and pushing away the different classes neighbors.
As for node distance function $d(h_1, h_2)$, where two node features $h_1, h_2 \in \mathbb{R}^{d}$, we compute 
the cosine distance between each node pair:
\begin{equation}
d(h_1, h_2)= 1 - \frac{h_1 \cdot h_2}{|h_1| \cdot |h_2|},
\end{equation}
where cosine distance is not affected by the absolute
value of the node vector. 

Define the samples with high prediction score on the minority classes as “anchor” samples.
Then we start from each anchor's middle representation $h_{a,j}$ (we use the hidden representation of
the final layer) of attribute label $j$ and take its $1$-hop neighbors
as positive samples $h_{+,j}$ or negative samples $h_{-,j}$ to construct neighbor-based triplet loss pairs.
The correlative loss function is defined as following:
\begin{equation}
L_{NTL} = \frac{\sum_{T}max(0, m_j + d(h_{a,j}, h_{+,j}) - d(h_{a,j}, h_{-,j}))}{|T|},
\end{equation}
where $h_{+,j}$ and $h{-,j}$ represent positive and negative samples with high confidence in the neighbors of central anchor respectively. Specifically, judged by two threshold hyper-parameters $\alpha_{+}(l)$ and $\alpha_{-}(l)$ of prediction scores, proposed model assign pseudo labels for positive or negative samples and selected anchors.

As shown in Figure~\ref{model}, we select high confidence minority class nodes as anchors and regularize the relative distances, which pulls the hard positive samples closer and pushes hard negative samples further. The number of positives, negatives and minority class anchors to be selected is determined by the the loss curriculum function $\alpha(l)$, where $l$ refers to current training epoch. Our proposed method can effectively deal with the over-smoothing problem of minority class nodes by pulling all the samples to well-classified side. 

\subsection{Curriculum Learning Framework}
In order to effectively solve the imbalanced issue on the graph, we first explore the suitable data generating strategy and design a special edge generator based on homophily and smoothness, and then define the classification loss $L_{GCL}$. Next, we propose a special metric loss $L_{NTL}$ according to the type relationship between the target node and its neighbor nodes with the help of pseudo labels to improve the quality of the generated nodes.
The final objective function is as follows:
\begin{equation}
\mathop{min}\limits_{\delta,\alpha,\lambda}L_{GCL} + \gamma \cdot L_{NTL}.
\end{equation}
The idea of curriculum learning\cite{26} demonstrates that learning from easy to hard significantly improves the generalization of the deep model. In order to leverage the training process, we design two contrary curriculum schedulers for loss functions:

The first one is the curriculum probability scheduler $\delta(l)$, which helps define sampling scale in one batch and makes data distribution from imbalance to balance.
This scheduler determines the sampling strategy for the proposed graph classification loss (GCL) function, where $L$ refers to expected total training epochs:
\begin{equation}
\delta(l) = \mu \cdot (1 - cos(\frac{l}{L} \cdot \frac{\pi}{2})).
\end{equation}
Here $\mu$ is the upper bound of sampling probability ranging from 0 to 1. The second one is the curriculum loss scheduler $\alpha(l)$, which controls the thresholds for judging anchors, positives and negatives for the neighbor-based triplet loss (NTL). Particularly for imbalanced data learning, what we want
is that the model first learns an appropriate feature representation in order to promote synthetic samples and benefit the classification. So that we hope the proposed model can assign more accurate pseudo labels in the training process with the following scheduler:
\begin{equation}
\alpha_+(l) = (1 - \beta_+ \cdot cos(\frac{l}{L} \cdot \frac{\pi}{2})),
\end{equation}
\begin{equation}
\alpha_-(l) =\beta_- \cdot cos(\frac{l}{L} \cdot \frac{\pi}{2}).
\end{equation}
In the early stage of training, metric loss occupies a larger proportion. On the one hand, it plays the role of "teacher" to guide the high-quality soft features and speed up the training process. On the other hand, it can help ensure better oversampling quality. In the later stage, system emphasizes more on the classification loss to learn the optimized classifier. 

\section{Experiments}
In this section, we design several experiments on five real-world datasets to verify the effect of GCN-CL. Three questions are solved in the followings:  
\begin{itemize}
\item[-] \textbf{RQ1:}
How is the performance of GCN-CL compared with the existing SOTA imbalanced classification methods?
\item[-] \textbf{RQ2:}
How do the GCL and NTL losses affect the classifier performance?
\item[-] \textbf{RQ3:}
How do different factors (imbalance ratio, sampling scale, base model, etc.) significantly affect the results of GCN-CL?
\end{itemize}

\subsection{Experimental Settings}
\subsubsection{Datasets}
\begin{table*}[htpb]
\centering
\resizebox{1.0\textwidth}{!}{
\begin{tabular}{|l||c|c|c|c|c|c|c|c|c|c|}
\hline
\diagbox{Model}{Dataset} &
 \multicolumn{2}{c|}{\textbf{Cora}} &
  \multicolumn{2}{c|}{\textbf{Citeseer}} &
 \multicolumn{2}{c|}{\textbf{BlogCategory}} &
 \multicolumn{2}{c|}{\textbf{Amazon Comp.}} &
 \multicolumn{2}{c|}{\textbf{Coauthor CS}} \\
 \hline
  \textbf{Metric}&
  cmA. &
  AUC-ROC &
  cmA. &
  AUC-ROC &
  cmA. &
  AUC-ROC &
  cmA. &
  AUC-ROC &
  cmA. &
  AUC-ROC \\
   \hline
  Origin&
  0.655$\pm$0.003 &
  0.902$\pm$0.005 &
  0.616$\pm$0.009 &
  0.883$\pm$0.002 &
  \underline{0.062$\pm$0.005} &
  0.569$\pm$0.009 &
  0.794$\pm$0.013 &
  0.980$\pm$0.002 &
  0.854$\pm$0.003 &
  0.977$\pm$0.002 \\
  Oversampling&
  0.645$\pm$0.025 &
  0.900$\pm$0.012 &
  0.619$\pm$0.011 &
  0.885$\pm$0.005 &
  0.056$\pm$0.002 &
  0.563$\pm$0.018 &
  0.798$\pm$0.002 &
  \underline{0.980$\pm$0.001} &
  0.853$\pm$0.006 &
  0.985$\pm$0.003 \\
  Reweighting&
  0.651$\pm$0.019 &
  0.909$\pm$0.009 &
  0.625$\pm$0.004 &
  \underline{0.886$\pm$0.001} &
  0.058$\pm$0.003 &
  0.561$\pm$0.017 &
  0.791$\pm$0.007 &
  0.978$\pm$0.001 &
  0.856$\pm$0.004 &
  0.980$\pm$0.002 \\
  DOS.&
  0.651$\pm$0.012 &
  0.901$\pm$0.006 &
  0.595$\pm$0.015 &
  0.875$\pm$0.005 &
  0.056$\pm$0.001 &
  0.556$\pm$0.011 &
  0.781$\pm$0.022 &
  0.977$\pm$0.003 &
  0.850$\pm$0.004 &
  0.976$\pm$0.002 \\
  GraphSMOTE&
  0.723$\pm$0.015 &
  0.915$\pm$0.007 &
  0.593$\pm$0.009 &
  0.870$\pm$0.007 &
  0.058$\pm$0.008 &
  0.558$\pm$0.005 &
  \underline{0.801$\pm$0.004} &
  0.978$\pm$0.001 &
  0.845$\pm$0.006 &
  0.976$\pm$0.002 \\
  \hline
  GNN-CL&
  \textbf{0.742$\pm$0.006} &
  \textbf{0.936$\pm$0.002} &
  \textbf{0.631$\pm$0.005}&
  \textbf{0.889$\pm$0.005}&
  \textbf{0.064$\pm$0.006}&
  \textbf{0.575$\pm$0.010}&
  \textbf{0.806$\pm$0.005}&
  \textbf{0.980$\pm$0.001}&
  \textbf{0.869$\pm$0.006}&
  \textbf{0.989$\pm$0.001}\\
  GNN-CL$_O$&
  0.669$\pm$0.018&
  0.911$\pm$0.007&
  0.627$\pm$0.011&
  0.884$\pm$0.006&
  0.052$\pm$0.001&
  0.561$\pm$0.011&
  0.799$\pm$0.007&
  0.979$\pm$0.001&
  0.862$\pm$0.007&
  0.988$\pm$0.001\\
  GNN-CL$_M$&
  \underline{0.725$\pm$0.016}&
  \underline{0.935$\pm$0.004}&
  0.617$\pm$0.006&
  0.883$\pm$0.002&
  0.055$\pm$0.004&
  \underline{0.569$\pm$0.004}&
  0.798$\pm$0.001&
  0.979$\pm$0.001&
  \underline{0.863$\pm$0.003}&
  \underline{0.989$\pm$0.001}\\
  GNN-CL$_C$&
  0.710$\pm$0.009&
  0.920$\pm$0.010&
  \underline{0.627$\pm$0.006}&
  0.881$\pm$0.001&
  0.059$\pm$0.003&
  0.565$\pm$0.007&
  0.791$\pm$0.004&
  0.977$\pm$0.002&
  0.858$\pm$0.010&
  0.985$\pm$0.002\\
   \hline
\end{tabular}
}
\caption{Experiment results for the imbalanced node classification task. (bold: best, underline: runner-up)}
\label{table1}
\end{table*}
For our experiments, we select 5 widely used node classification datasets belonging to 4 types for experimentation comparison, including two well-known citation graphs Citeseer and Cora\cite{26}, Co-purchase graph: Amazon computers\cite{23}, Co-authorship graph: Coauthor CS\cite{23} and Co-authorship graph: BlogCatalog\cite{27}. The detailed introduction of these datasets is placed in Appendix B.1.

\subsubsection{Compared methods}
We compare GNN-CL with representative
and state-of-the-art approaches for handling imbalanced class distribution problem, which includes conventional methods: Oversampling and Reweighting, deep learning method Deep OverSampling and graph neural network method GraphSMOTE. Similarly, due to the length limitation, we give detailed descriptions of these baselines in Appendix B.2. 
In order to verify the effectiveness of each part of our proposed method, four variants including ablation models of GraphSMOTE are tested:
\begin{itemize}
\item[-] \textbf{GNN-CL}
Our proposed graph neural network with curriculum learning on metric loss and classification loss.
\item[-] \textbf{GNN-CL$_O$}
It removes the oversampling strategy from proposed model, so that synthetic nodes and corresponding edges will not be generated.
\item[-] \textbf{GNN-CL$_M$}
It removes the metric loss part from proposed model and ignores the regularization between neighbors.
\item[-] \textbf{GNN-CL$_C$}
It removes the curriculum learning mechanism from proposed model, and the ratio of two losses is determined by the fixed experimental optimal parameters.
\end{itemize}

\subsubsection{Metrics}
In order to comprehensively measure the effect of our proposed model, we adopt three commonly used imbalance classification task criterias: class balanced mean accuracy (cmA) and mean AUR-ROC score. cmA is computed on all testing examples at once, Following the standard profile, we apply the class-balanced accuracy defined as the average of recall obtained on each class. It can be formulated as following:
\begin{equation}
cmA = \frac{\sum_{i=1}^{|C|}\frac{TP_i}{TP_i+FN_i}}{|C|}.
\end{equation} 
AUC-ROC score indicates the probability that the predicted positive case is ranked higher than other classes.

\subsection{Overall Performance (RQ1)}
Here we compare the effectiveness of different methods by the imbalanced semi-supervised node classification task on various datasets. In order to eliminate variance, we repeat the process for 5 times and report the averaged cmA. and AUC-ROC in TABLE~\ref{table1}. As we can see, GNN-CL achieves the best and stable performance. Except GNN-CL, the performances have ups and downs on the classic baselines (Oversampling, Reweighting, Dos.). In general, the effect of these methods is similar to that of origin, which shows that the traditional methods are not suitable for graph structure data. Graph-based GraphSMOTE method has certain superiority in some datasets with small imbalance ratio, such as Cora and Amazon comp., but performs poorly in other datasets.

According to the ablation experiment, the performance of removing oversampling module is significantly weakened. Howerver there still have some good cases on some datasets when omiting metric loss, which shows that sampling is more important. Metric learning mainly assists the classification task by improving the quality of node representations. Our GNN-CL has around $1-3\%$ performance gain over the best baseline in general, which indicates that oversampling and metric learning modules alleviate the adverse effects of long-tail distribution.

\subsection{Process Analyses (RQ2)}
\begin{figure}
\centering
\includegraphics[width=0.48\textwidth]{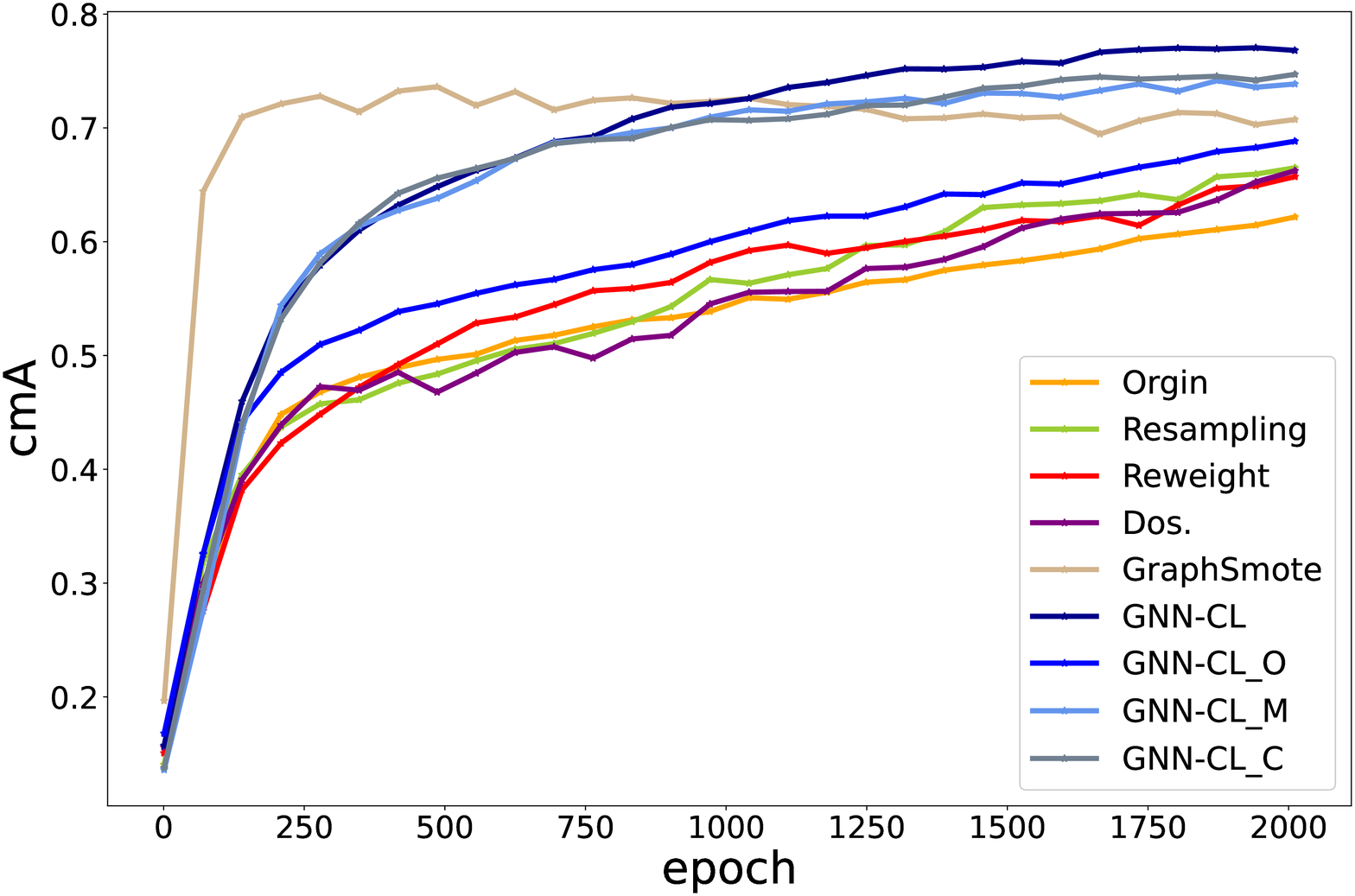}
\caption{Test results (cmA) of each method during training process on Cora dataset.}
\label{fig:status}
\end{figure} 
To further illustrate the effectiveness of GNN-CL, we draw a group of training process curves to make comparisons intuitively. In Figure~\ref{fig:status}, we draw the cmA scores of each comparison method, where our proposed method and its variants in the training process with different colors. From this visualization, we can see that proposed GNN-CL model achieves the best classification results as well as a stable training process. Other traditional sampling methods are better than the origin method, which shows that the general sampling methods are also applicable on the graph.

However, compared with the special graph sampling method, it has obvious disadvantages in performance, which shows that GraphSMOTE and GNN-CL have good effects when applied in imbalance classification situations on graphs. It is worth noting that GraphSMOTE needs fewer rounds to reach its peak and then begins to decline, proving that the generation of nodes and edges is obviously helpful at the beginning of training. But because sampling process remains unchanged, over-fitting problem comes soon and models can not get the best result.

Our GNN-CL model gives priority to the quality of soft features at the beginning of training through curriculum learning mechanism, and then turns to generate high confidence nodes and edges gradually. Compared with baseline methods, it not only ensures the speed of training, but also improves the effect continuously and significantly. Ablation models GNN-CL$_M$ and GNN-CL$_C$ also have comparatively good results, but due to the lack of metric loss or curriculum learning, the effect falls behind significantly in the later stage of training.

\subsection{In-depth Analysis (RQ3)}
\subsubsection{Study on Imbalance Ratio}
\begin{table}[htpb]
\centering
\resizebox{0.45\textwidth}{!}{
\begin{tabular}{|l||c|c|c|c|c|}
\hline
&
 \multicolumn{5}{c|}{\textbf{Imbalance Ratio}}\\
 \hline
  Methods&
  0.1 &
  0.3 &
  0.5 &
  0.7 &
  0.9  \\
 \hline
  Origin&
  0.354 &
  0.554 &
  0.634 &
  0.677 &
  0.711  \\
  Oversampling&
  0.442 &
  0.559 &
  0.659 &
  0.676 &
  0.713 \\
  Reweighting&
  0.464 &
  0.579 &
  0.661 &
  0.699 &
  0.702 \\
  DOS.&
  0.474 &
  0.587 &
  0.644 &
  0.625 &
  0.633 \\
  GraphSMOTE&
  \textbf{0.598} &
  0.708 &
  0.723 &
  0.731 &
  \underline{0.755} \\
  \hline
  GNN-CL&
  0.591&
  \underline{0.712}&
  \textbf{0.745}&
  \textbf{0.757}&
  \textbf{0.759}\\
  GNN-CL$_O$&
  0.473&
  0.582&
  0.670&
  0.714&
  0.721\\
  GNN-CL$_M$&
  \underline{0.597}&
  \textbf{0.724}&
  \underline{0.740}&
  \underline{0.745}&
  0.746\\
  GNN-CL$_C$&
  0.542&
  0.683&
  0.701&
  0.734&
  0.745\\
   \hline
\end{tabular}
}
\caption{Experiment results on different imbalance ratio. (bold: best, underline: runner-up)}
 \label{table:ir}
\end{table}
The classification performances of all above models under different imbalance ratios are listed in Table~\ref{table:ir}. The severity of the imbalance problem is in reverse proportion to the value of imbalance ratio.
It can be seen that the two graph-based sampling methods are significantly effective, especially when the imbalance ratio value is small.
For example, when imbalance ratio $= 0.1$, there exists an increase of more than $20\%$ compared with origin. But in this extremity, there is no obvious distinction between GraphSMOTE and GNN-CL, indicating that due to the serious imbalance problem, the metric learning module is not fully be used. Overall, it can be seen that sampling plays a greater role than metric learning on the Cora dataset. When imbalance ratio $= 0.9$, the dataset is basically in balance, so that the sampling methods have little significance.
\subsubsection{Study on Curriculum Learning Rate}
\begin{figure}[tb]
  \centering
  \subfigure[GCN]{\label{fig1:1}
  \begin{minipage}[t]{0.22\textwidth}
  \centering
  \includegraphics[width=1\textwidth]{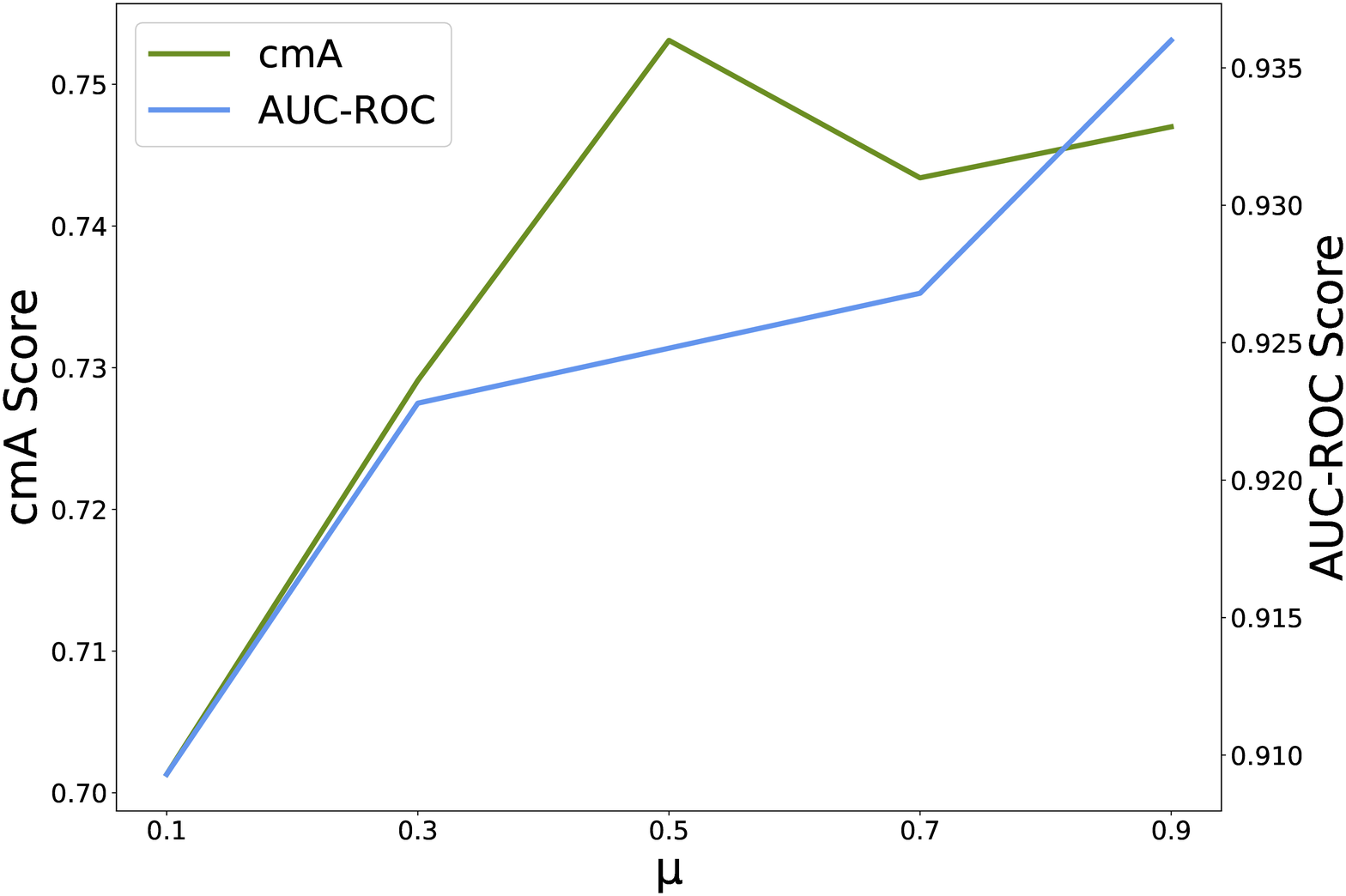}
  \end{minipage}
  }
  \subfigure[GraphSAGE]{\label{fig1:2}
  \begin{minipage}[t]{0.22\textwidth}
  \centering
  \includegraphics[width=1\textwidth]{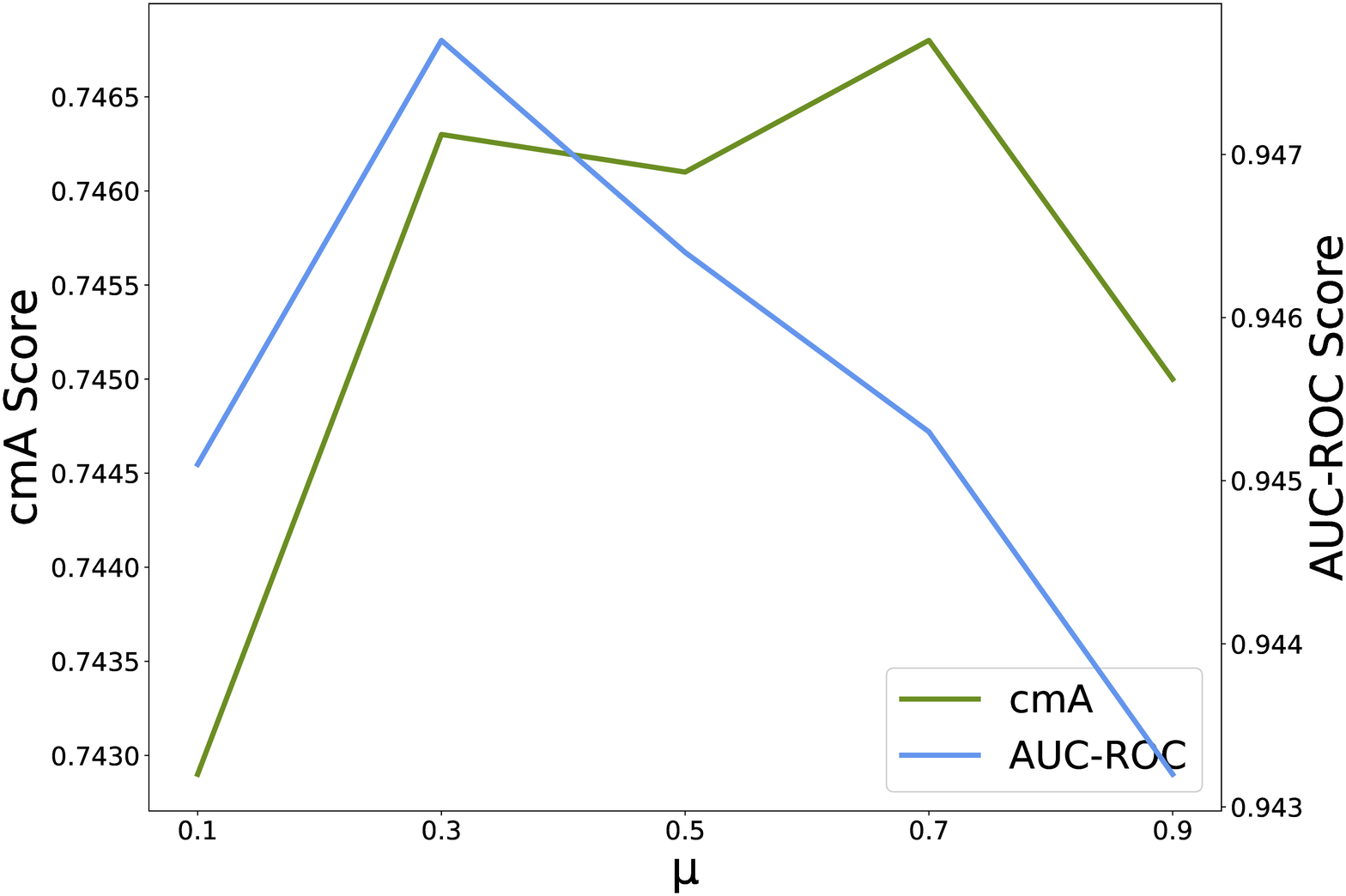}
  \end{minipage}
  }
  \caption{Experiment results on diiferent oversampling scale. }
  \label{fig:oversampling}
\end{figure}

\begin{figure}[tb]
  \centering
  \subfigure[Decision boundary of positive label]{\label{fig1:1}
  \begin{minipage}[t]{0.22\textwidth}
  \centering
  \includegraphics[width=1\textwidth]{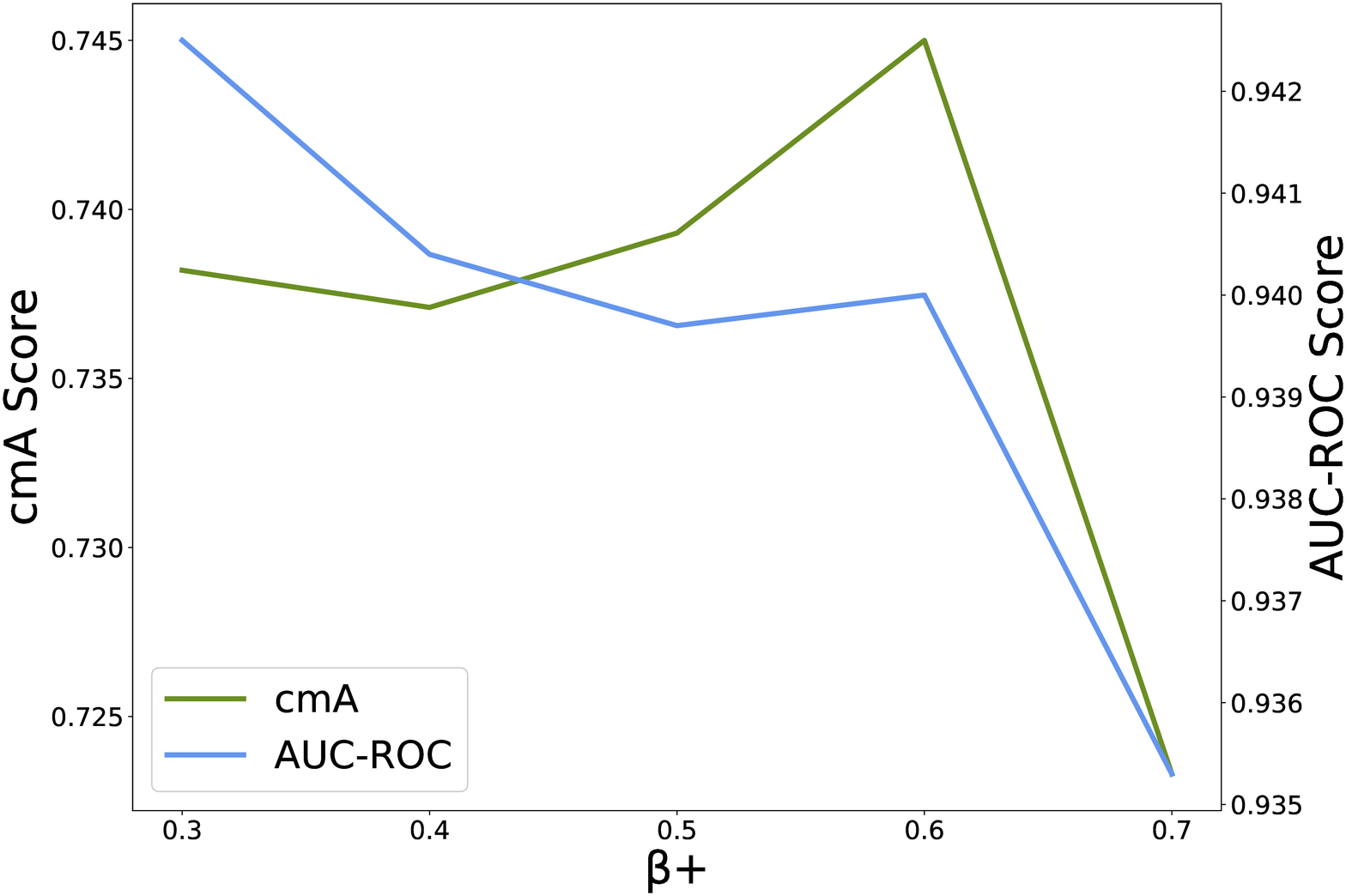}
  \end{minipage}
  }
  \subfigure[Decision boundary of negative label]{\label{fig1:2}
  \begin{minipage}[t]{0.22\textwidth}
  \centering
  \includegraphics[width=1\textwidth]{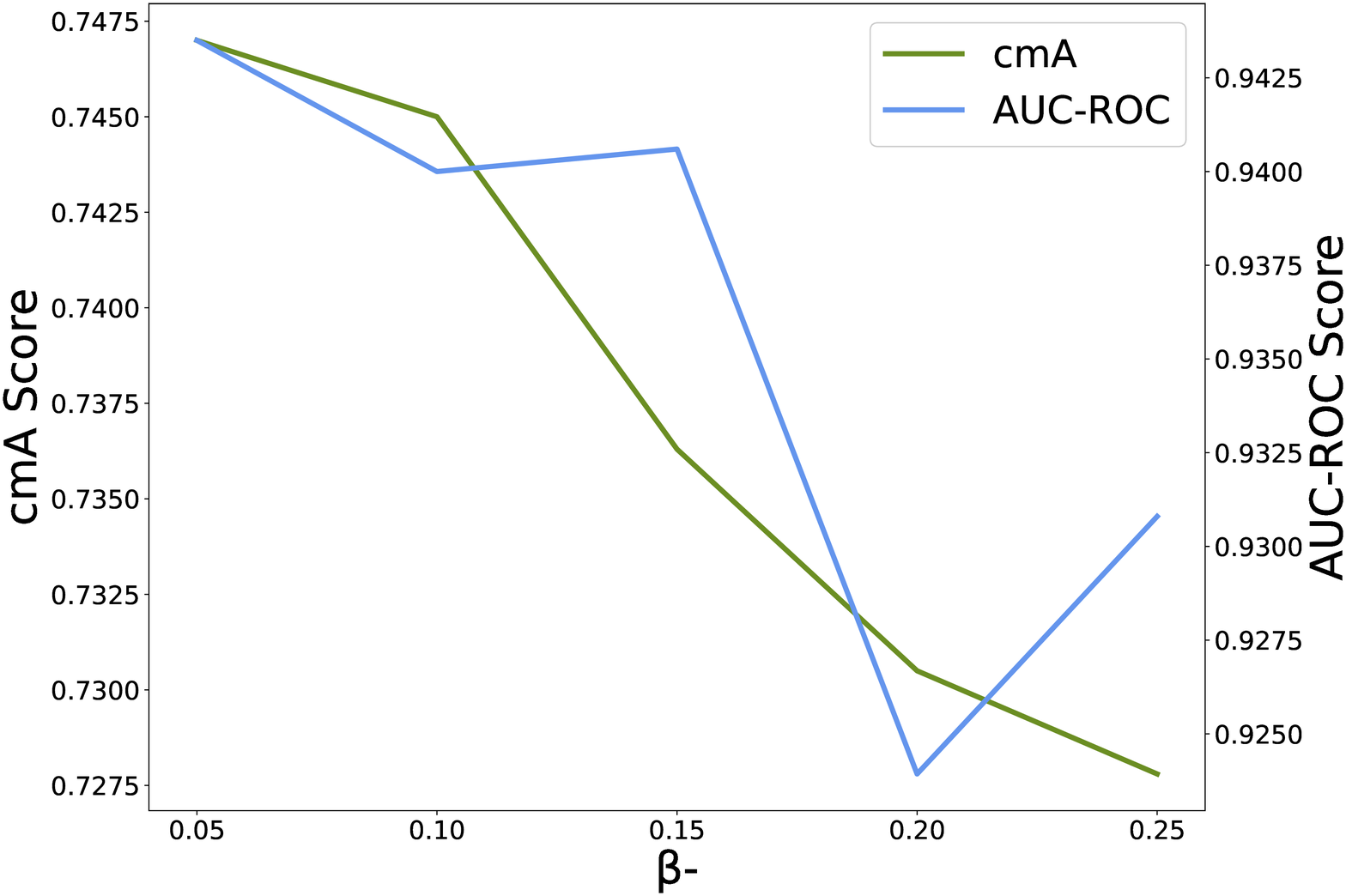}
  \end{minipage}
  }
  \caption{Experiment results on metric learning scale. }
  \label{fig:metric}
\end{figure}

In this section, we verify the impact of different rates for the curriculum learning mechanism on the results, shown in Figure~\ref{fig:oversampling} and Figure~\ref{fig:metric}. 1) Firstly, we use GCN and GraphSAGE base models to test the hyper-parameter $\mu$, which controls the upper bound of the probability related to the sampling scale. It can be seen from Figure~\ref{fig:metric}(a) that generating more synthetic nodes on the Cora helps to improve the performance of the model. Because Cora has relatively small size and slight imbalance problem, the quality of generated nodes is high. In spatial domain method such as GraphSAGE, appropriate sampling scale can achieve better results. 2) We also test the parameters used to judge pseudo labels in metric learning module $\beta$. $\beta_+$ and $\beta_-$ control the possibility of generating positive and negative sample labels. It can be seen from the Figure~\ref{fig:metric}(b) that too many pseudo labels are not conducive to the clear classification boundary. The best model performance can be obtained only with appropriate parameters. Experiments on other hyper-parameters are in Appendix B.4.

\subsubsection{Study on Base Model}
\begin{table}[htpb]
\centering
\resizebox{0.5\textwidth}{!}{
\begin{tabular}{|l||c|c||c|c|}
\hline
 &
 \multicolumn{2}{c||}{\textbf{Cora}}&
 \multicolumn{2}{c|}{\textbf{Citeseer}}\\
 \hline
  Metrics&
  mcA &
  AUC-ROC &
  mcA &
  AUC-ROC  \\
 \hline
  Origin&
  0.681$\pm$0.024 &
  0.907$\pm$0.005 &
  0.620$\pm$0.020 &
  0.859$\pm$0.010  \\
  Oversampling&
  0.663$\pm$0.029 &
  \underline{0.914$\pm$0.012} &
  0.621$\pm$0.006 &
  0.864$\pm$0.002 \\
  Reweighting&
  0.675$\pm$0.005 &
  0.904$\pm$0.004 &
  0.636$\pm$0.008 &
  0.867$\pm$0.004 \\
  DOS.&
  0.689$\pm$0.010 &
  0.908$\pm$0.008 &
  0.609$\pm$0.011 &
  0.852$\pm$0.006 \\
  GraphSMOTE&
  0.673$\pm$0.008 &
  0.905$\pm$0.002 &
  0.605$\pm$0.009&
  0.852$\pm$0.002 \\
  \hline
  GNN-CL&
  \textbf{0.703$\pm$0.007}&
  0.911$\pm$0.007&
  \textbf{0.646$\pm$0.004}&
  \textbf{0.881$\pm$0.004}\\
  GNN-CL$_O$&
  0.686$\pm$0.021&
  \textbf{0.916$\pm$0.008}&
  \underline{0.636$\pm$0.005}&
  \underline{0.873$\pm$0.002}\\
  GNN-CL$_M$&
  0.693$\pm$0.003&
  0.909$\pm$0.006&
  0.618$\pm$0.002&
  0.858$\pm$0.005\\
  GNN-CL$_C$&
  \underline{0.698$\pm$0.001}&
  0.910$\pm$0.004&
  0.625$\pm$0.005&
  0.857$\pm$0.004\\
   \hline
\end{tabular}
}
\caption{Experiment results on different base models. (bold: best, underline: runner-up)}
\label{table:gcn}
\end{table}
As shown in Table~\ref{table:gcn}, we try to apply proposed GNN-CL method to other base models to verify the generality. As for GraphSMOTE model, its ability to solve the imbalance problem is similar to that of other traditional methods, while GNN-CL model has consistent applicability on all datasets. The performance of GNN-CL is $2-4\%$ higher than the optimal baseline. It can be seen from the ablation models that oversampling and metric learning modules have their own advantages.

\section{Conclusion}
In this paper, we mainly focus on processing the imbalance problem in complex node classification task and give a novel graph neural network framework with curriculum learning (GNN-CL). On this foundation, adaptive graph oversampling and neighbor-based metric learning are proposed for dynamic sampling operation and loss backward propagation. Extensive experiments prove that our final proposed approach GNN-CL outperforms state-of-the-art methods in different areas with consistent level of performance. In the future, we will explore other possible graph sampling methods and design an interpretable end-to-end learning framework.

\newpage
\appendix
\section{Homophily and Model Performance}
\begin{figure*}[h]
  \centering
  \subfigure[Cora]{\label{fig:homo:1}
  \begin{minipage}[t]{0.18\textwidth}
  \centering
  \includegraphics[width=1\textwidth]{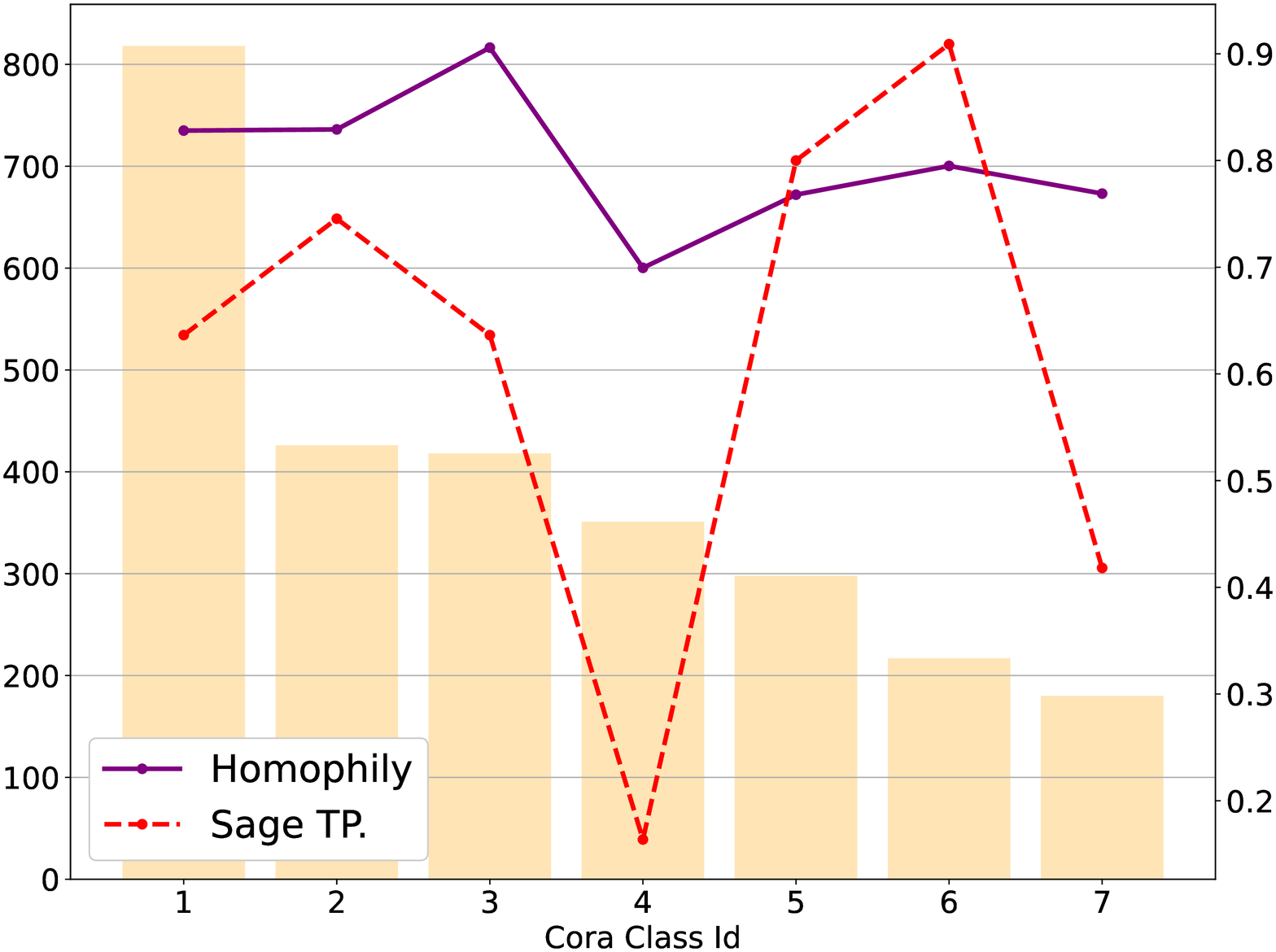}
  \end{minipage}
  }
  \subfigure[Citeseer]{\label{fig:homo:2}
  \begin{minipage}[t]{0.18\textwidth}
  \centering
  \includegraphics[width=1\textwidth]{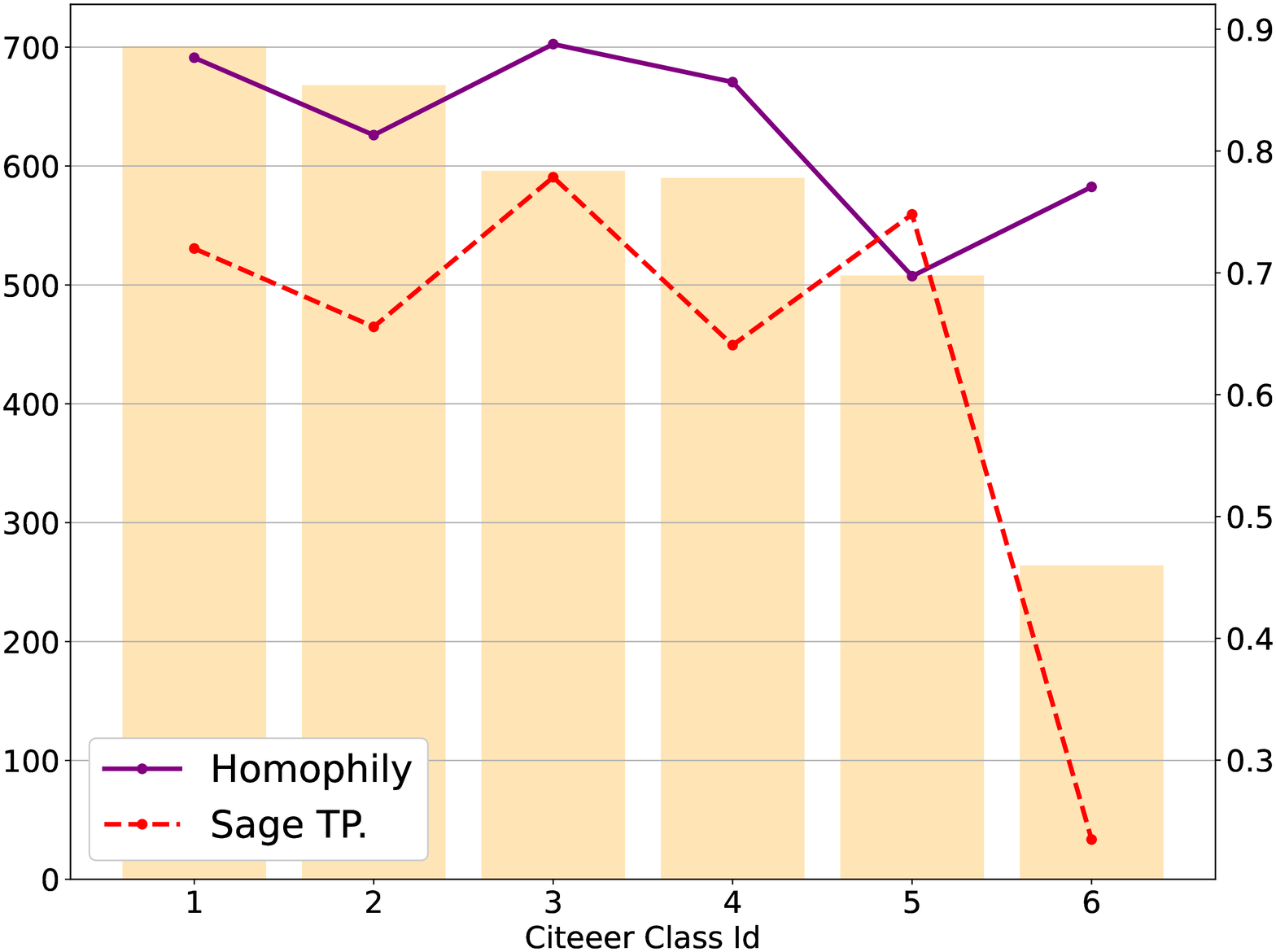}
  \end{minipage}
  }
  \subfigure[BlogCategory]{\label{fig:homo:3}
  \begin{minipage}[t]{0.18\textwidth}
  \centering
  \includegraphics[width=1\textwidth]{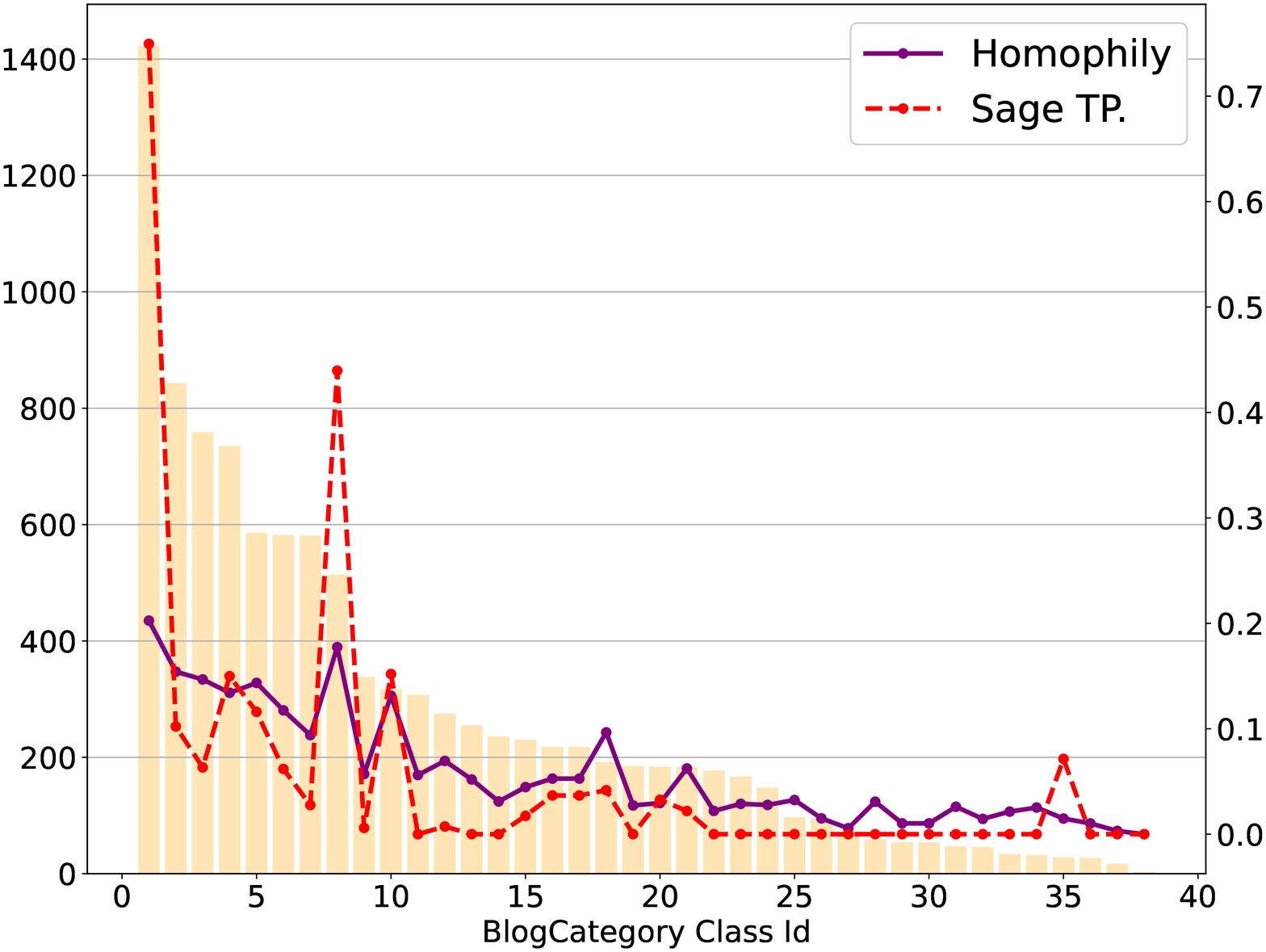}
  \end{minipage}
  }
  \subfigure[Amazon Comp.]{\label{fig:homo:4}
  \begin{minipage}[t]{0.18\textwidth}
  \centering
  \includegraphics[width=1\textwidth]{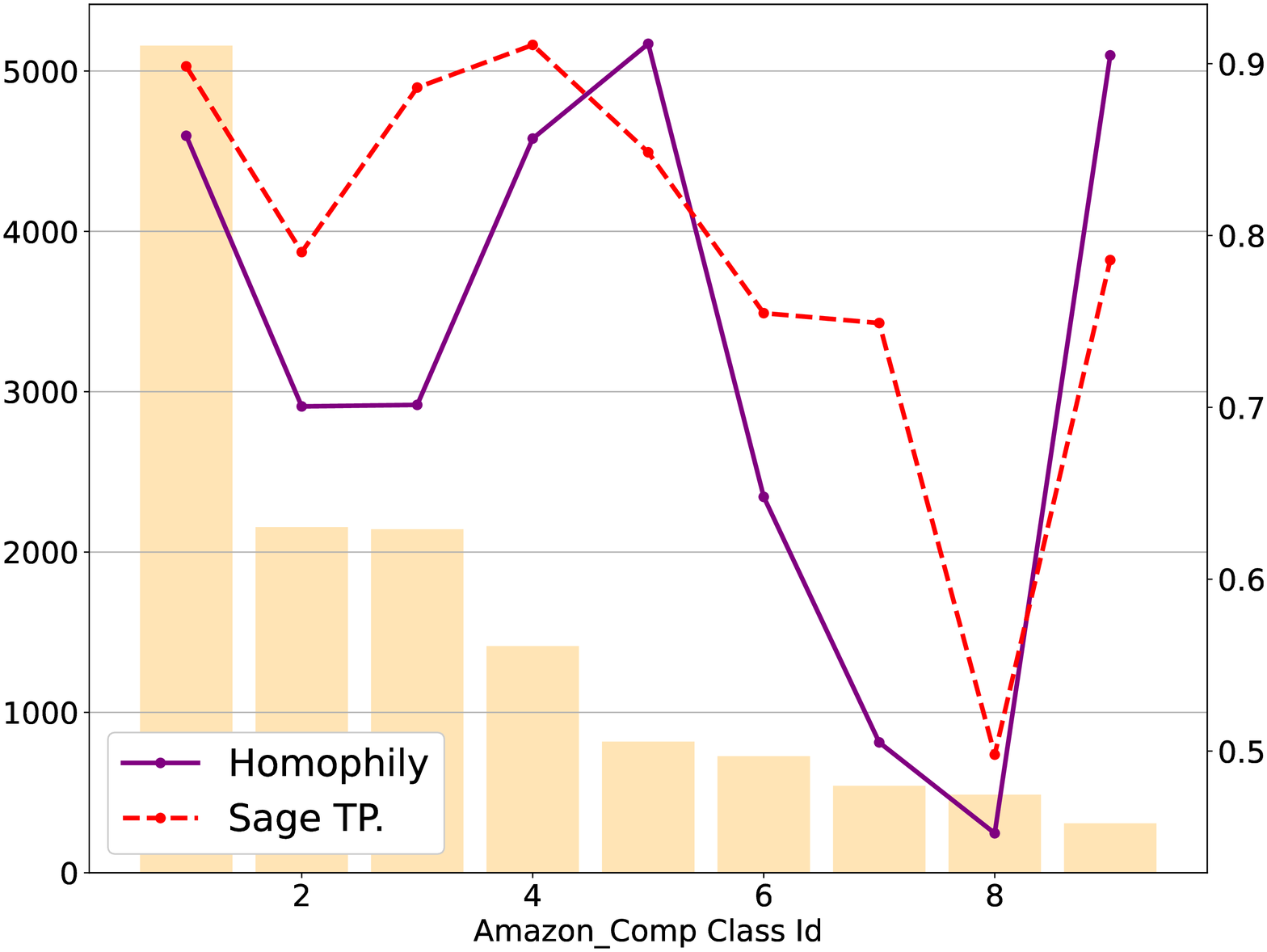}
  \end{minipage}
  }
  \subfigure[Coauthor CS]{\label{fig:homo:5}
  \begin{minipage}[t]{0.18\textwidth}
  \centering
  \includegraphics[width=1\textwidth]{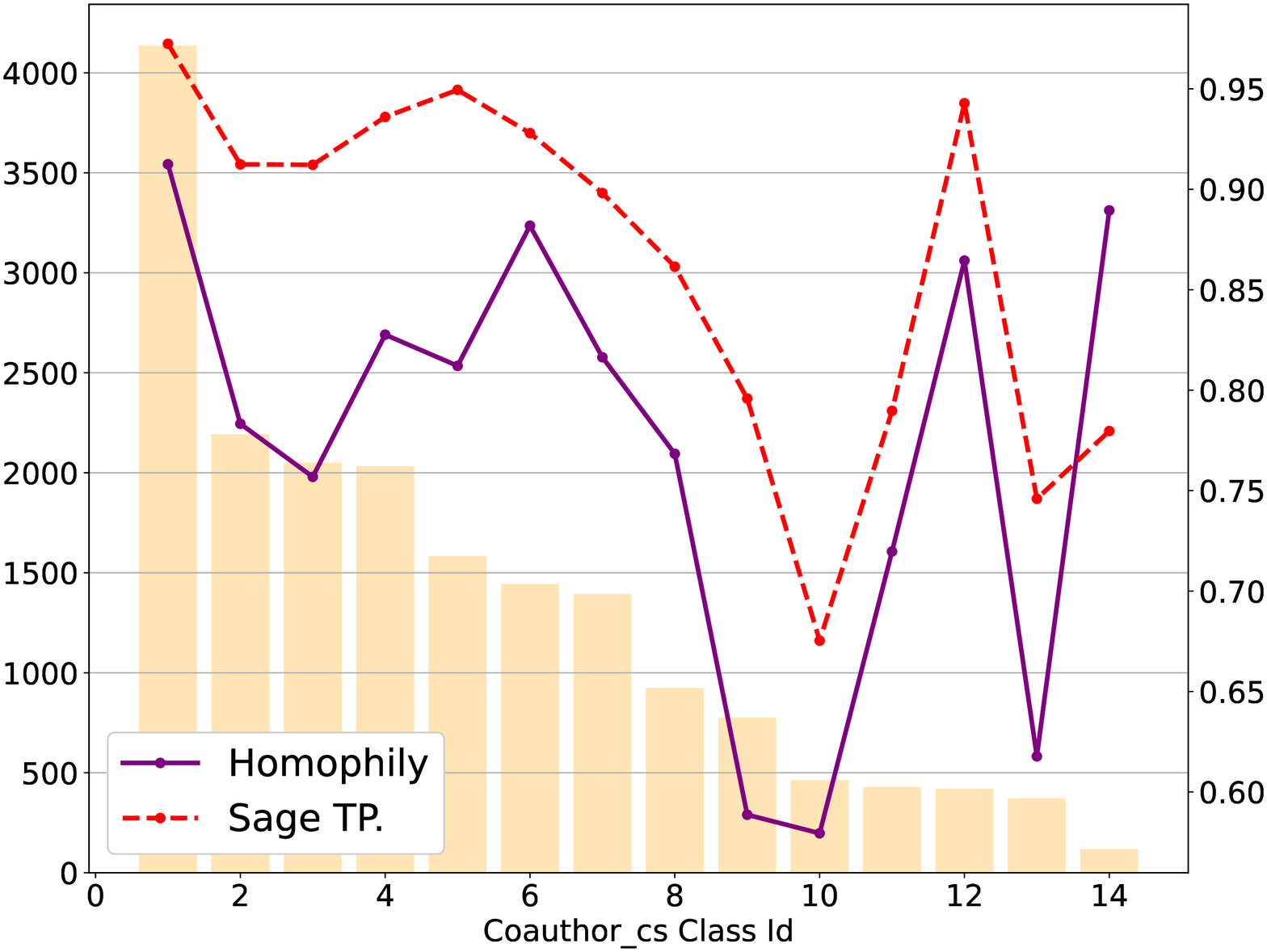}
  \end{minipage}
  }
  \caption{Homophily score (purple) and True Positive (TP.) score (red) via GraphSAGE model of different class in each dataset.}
  \label{fig:homo}
\end{figure*}

In Figure~\ref{fig:homo}, we display the GraphSAGE model performance True Positive (TP.) in different classes and the corresponding homophily score of each class sets in five dataset. The histogram represents the number of points in each class from high to low. We can find that in these five datasets, when homophily value decreases, the representation quality of GNN model will be lessened, especially for tail class in Figure~\ref{fig:homo}(c)(d)(e). 
However, the decline of the performance is not only influenced by homophily. For example, in Cora and Citeseer, homophily value maintains at high level but the effect of tail nodes decreases because the classifier cannot be trained effectively.
\section{Supplement of Experiments}
\subsection{Details of datasets}
\begin{table*}[tb]
\caption{Statistics of the datasets.}
\begin{center}
\begin{tabular}{|c|c|c|c|c|c|c|c|}
\hline
\textbf{Dataset}&\textbf{$\sharp$Node}& \textbf{$\sharp$Edge}&\textbf{$\sharp$Training}&\textbf{$\sharp$Validation}&\textbf{$\sharp$Test}&\textbf{Imbalance Ratio(M:1)}\\
\hline
Cora&$2708$&$10556$&$140$&$140$&$2428$&$-$ \\
Citeseer&$3327$&$9228$&$831$&$831$&$1663$&$2.65$ \\
BlogCategory&$10312$&$667966$&$2561$&$2561$&$5146$&$355.00$ \\
Amazon Comp.&$13752$&$287209$&$3434$&$3434$&$6875$&$16.74$ \\
Coauthor CS&$18333$&$163788$&$4579$&$4579$&$9164$&$35.65$ \\
\hline
 \end{tabular}
\end{center}
\label{table:dataset}
\end{table*}
Here we list the statistics of the datasets in Table~\ref{table:dataset}.
1) We first use the two well-known citation network datasets: Cora and Citeseer. Edges in these networks represent the citation relationship between two papers (undirected), node features are the bag-of-words vector of the papers and labels are the fields of papers. Among them, Cora contains 140 labeled training nodes with balanced class distributions, so the factor $imbalance\_ratio$ is used to disequilibrate data by downsampling half of random classes. For each minority class, the number is 20$\times imbalance\_ratio$. Meanwhile, there is a mild class imbalance problem in the training set of Citeseer.
2) Amazon computer is built from fragments in the Amazon co-purchase graph. The nodes in the graph represent products, and their features are obtained through the bag-of-words model of consumers' comments. The edges represent that the products are purchased at the same time and the category label is obtained by the category of the product. It contains 9 types of samples and the head majority class is 16 times more than the tail minority class.
3) Coauthor CS is a co-authorship graph based on the Microsoft academic graph. The nodes symbolize the authors and the edges represent the co-authorship relationships. The features originate from paper keywords for each author's paper. At the same time, 14 different labels indicate most active fields of study for each author suffering from a large imbalance problem.
4) BlogCatalog is a co-authorship graph based on the Microsoft academic graph. The nodes symbolize the authors and the edges represent the co-authorship relationships. The features originate from paper keywords for each author's paper. At the same time, class labels indicate most active fields of study for each author. Classes in this dataset meet a genuine imbalanced distribution, with 14 classes smaller than 100, and 8 classes larger than 500.

\subsection{Details of Compared Methods}
The supplementary descriptions of the compared methods are as follows:
\begin{itemize}
\item[-] \textbf{Oversampling}
Oversampling is a classical method, which improves the performance of classifier by repeating minority classes. In the implementation, we duplicating $n_s$ minority samples and edges connected with them on the graph.
\item[-] \textbf{Reweighting}
This is a kind of method to adjust the category weight of loss function, mainly by increasing the importance of a few categories in supervision information.
\item[-] \textbf{Deep OverSampling}
To counteract class imbalance problem, this method utilize a synthetic embedding target in the deep feature space, which is sampled from the linear subspace of in-class neighbors.
\item[-] \textbf{GraphSMOTE}
GraphSMOTE\cite{25} synthesizes similar new samples in graph embedding space to assure genuineness. In addition, an edge generator is trained simultaneously to model the relation information, and provide it for those new samples.
\end{itemize}

\subsection{Settings and Hyper-parameters}
\begin{figure}[tb]
  \centering
  \subfigure[The parameter in graph classification loss.]{\label{fig1:1}
  \begin{minipage}[t]{0.22\textwidth}
  \centering
  \includegraphics[width=1\textwidth]{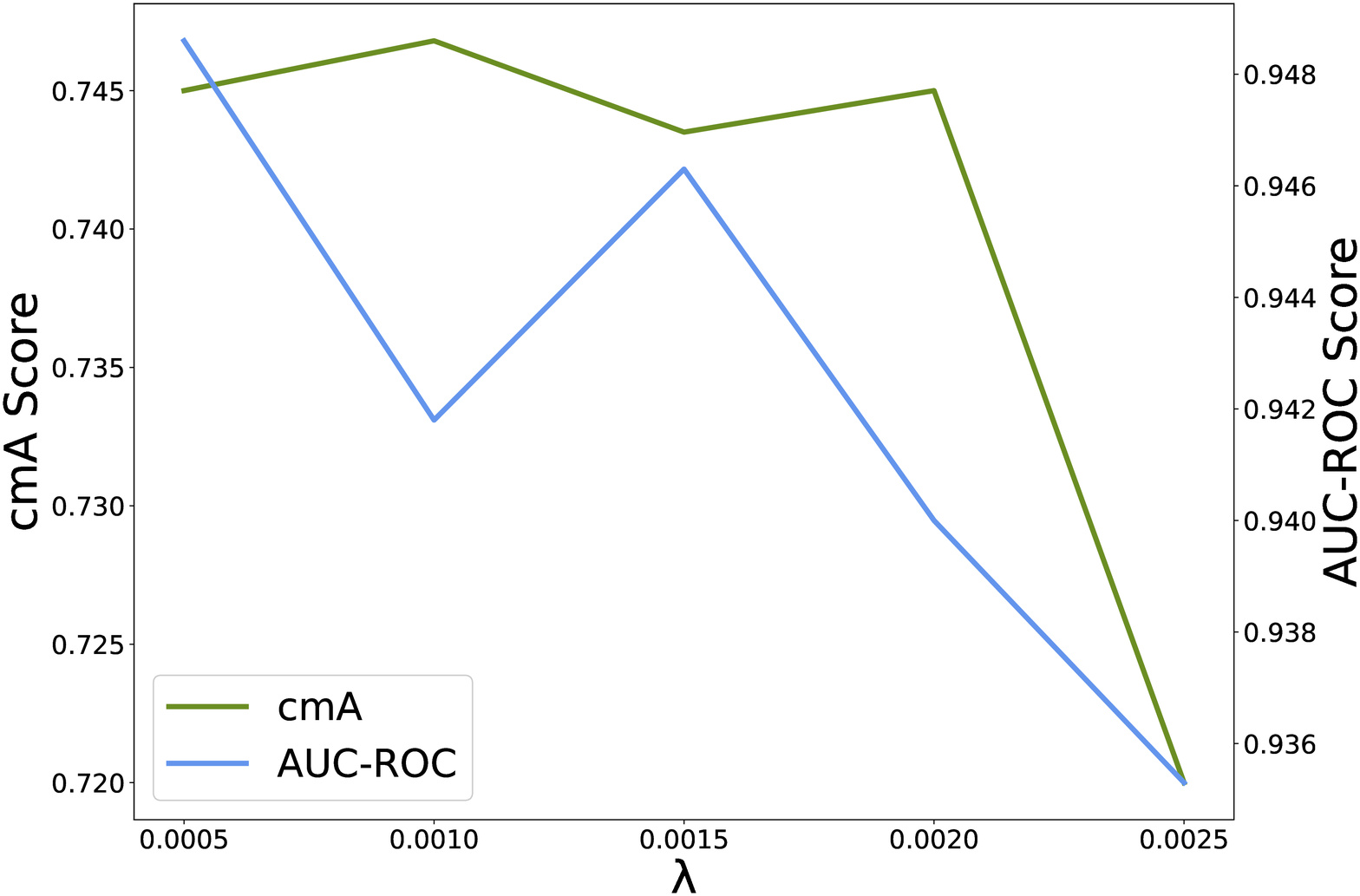}
  \end{minipage}
  }
  \subfigure[The parameter in final objective function.]{\label{fig1:2}
  \begin{minipage}[t]{0.22\textwidth}
  \centering
  \includegraphics[width=1\textwidth]{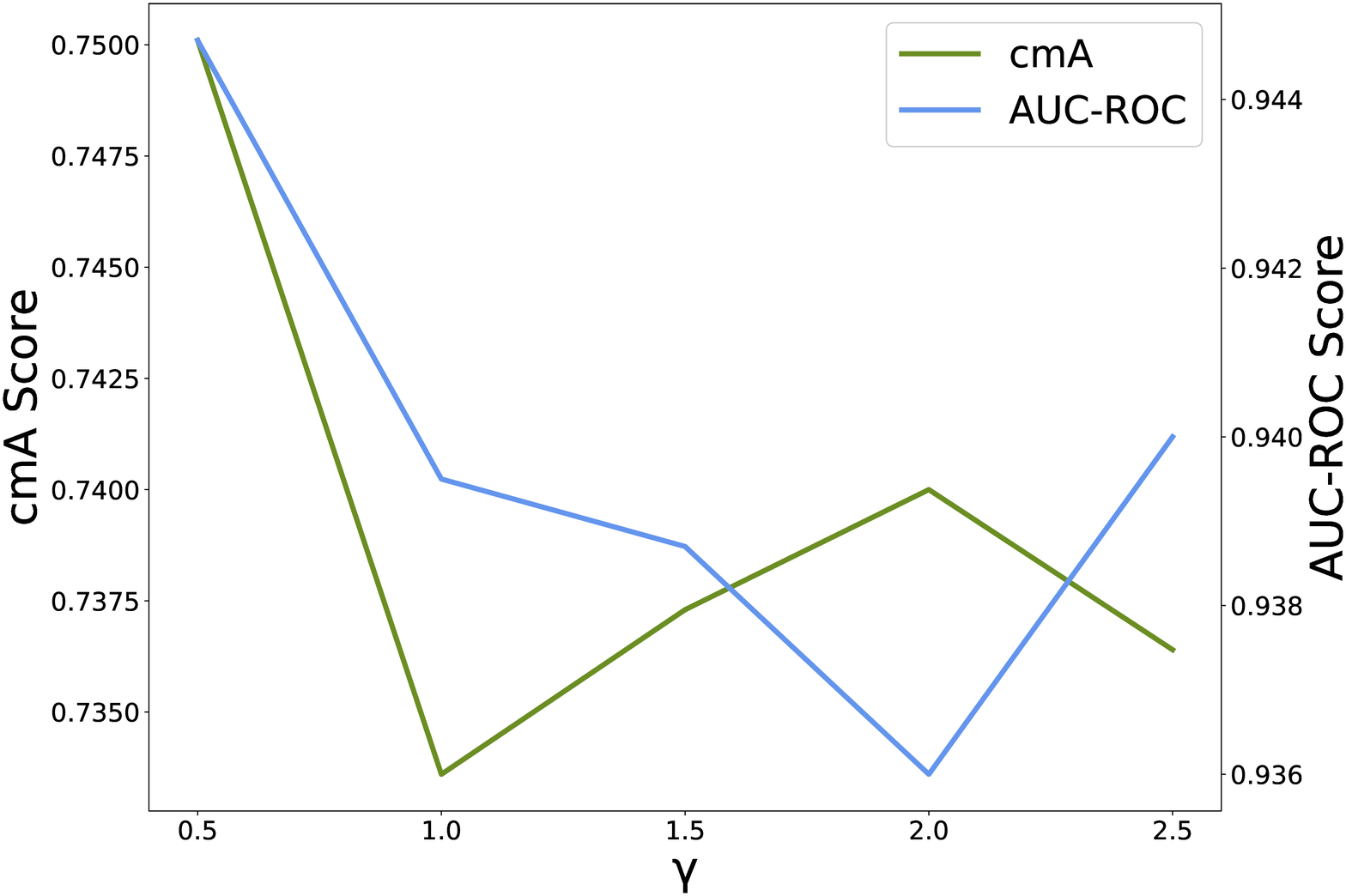}
  \end{minipage}
  }
  \caption{Experiment results on other hyper-parameters. }
  \label{fig:hyperparameters}
\end{figure}

For all methods in the experiment, we randomly initialize parameters and use Adam to optimize the model with a maximum of 2000 epochs (adopting early stopping with a patience of 100). In practice, we implement them with pytorch1.2 to train model parameters and also use mini-batch gradient descent, which divides training data into several batches and updates parameters by each batch. The learning rate in all mothods is initialized to 0.001 and the weight decay is set to 0.0005. Two hyper-parameters $\lambda$ and $\gamma$ are set to 0.002 and 1.0 by default, according to the actual function values in the function. Other experimental parameters, the oversampling parameter $\mu $ is set to 1.0, while the boundary parameter $\beta_+$ and $\beta_-$ of pseudo labels in $L_{NTL}$ are set to 0.6 and 0.1, respectively. $m$ in neighbor-based triplet loss is empirically set to 0.5. In addition, several sensitivity experiments are carried out to explore the proper range of parameters.

\subsection{Study On Other Parameters}
In this section,
we do sensitivity analysis to some essential parameters in
GNN-CL and Figure~\ref{fig:hyperparameters} shows the training process curves on Cora. 1) We first test the effect of the ratio of graph classification loss and edge generator loss, shown in Figure~\ref{fig:hyperparameters}(a). As the proportion of edge generator loss grows, the performance keeps steadily first and then shows a continued decline. The optimal performance is obtained
when $\lambda < 0.0015$. 2) We also investigate the effect of the ratio of classification loss and metric loss reported in Figure~\ref{fig:hyperparameters}(b). Based on the results, we can find that limiting $\gamma$ to a smaller range works best. In the future, we can adjust the proportional parameter by standardizing the loss function.
\end{document}